\documentclass{INTERSPEECH2023}
\usepackage{xurl}
\usepackage{multirow}
\usepackage{cite}           

\interspeechcameraready

\title{Comparison of Multilingual Self-Supervised and Weakly-Supervised Speech Pre-Training for Adaptation to Unseen Languages}
\name{Andrew Rouditchenko$^1$, Sameer Khurana$^1$, Samuel Thomas$^{2,3}$, Rogerio Feris$^{2,3}$, Leonid Karlinsky$^{2,3}$, Hilde Kuehne$^{3,4}$, David Harwath$^5$, Brian Kingsbury$^{2,3}$, James Glass$^1$}

\address{$^1$MIT, USA \hspace{2mm}
  $^2$IBM Research AI, USA \hspace{2mm}
  $^3$MIT-IBM Watson AI Lab, USA \hspace{2mm}
  $^4$Goethe University Frankfurt, Germany \hspace{2mm}
  $^5$UT Austin, USA}
\email{roudi@mit.edu}

\begin{document}

\maketitle
 
\begin{abstract}
Recent models such as XLS-R and Whisper have made multilingual speech technologies more accessible by pre-training on audio from around 100 spoken languages each.
However, there are thousands of spoken languages worldwide, and adapting to new languages is an important problem.
In this work, we aim to understand which model adapts better to languages unseen during pre-training.
We fine-tune both models on 13 unseen languages and 18 seen languages.
Our results show that the number of hours seen per language and language family during pre-training is predictive of how the models compare, despite the significant differences in the pre-training methods.
\end{abstract}

\section{Introduction}
Multilingual speech processing has seen rapid progress thanks to the availability of more data and larger models.
Models like XLS-R~\cite{babu2021xls} and Whisper~\cite{radford2022robust} are pre-trained on hundreds of thousands of hours of audio from around 100 spoken languages.
Downstream users can adapt these publicly available models for Automatic Speech Recognition (ASR) on different languages and domains~\cite{khurana2022magic,krishna2021using,zhao2022improving}.
XLS-R follows Wav2Vec2.0's~\cite{baevski2020wav2vec} self-supervised pre-training process, which only requires unlabeled speech, and it can be adapted for ASR with a small amount of labeled data.
In contrast, Whisper was pre-trained with labeled speech-text pairs acquired by scraping the web and can be directly applied to ASR in many languages without requiring fine-tuning.

Although these models are pre-trained on many languages, they still omit several thousand spoken languages during pre-training, which limits their applicability to these unseen languages.
Many of these unseen languages have over a million speakers, making them sufficiently widespread to be of importance.
We believe more audio in these languages will become available through projects like CommonVoice~\cite{ardila-etal-2020-common}.
However, it is infeasible to repeat pre-training each time more data is released since the compute required is only accessible to the largest corporations.
For example, XLS-R~\cite{babu2021xls} was pre-trained on 128 GPUs for smaller models and 200 GPUs for larger models.
Therefore, we find it essential to study the adaptation of these pre-trained models to unseen languages.

\begin{figure}[t]
    \centering
    \includegraphics[width=\linewidth]{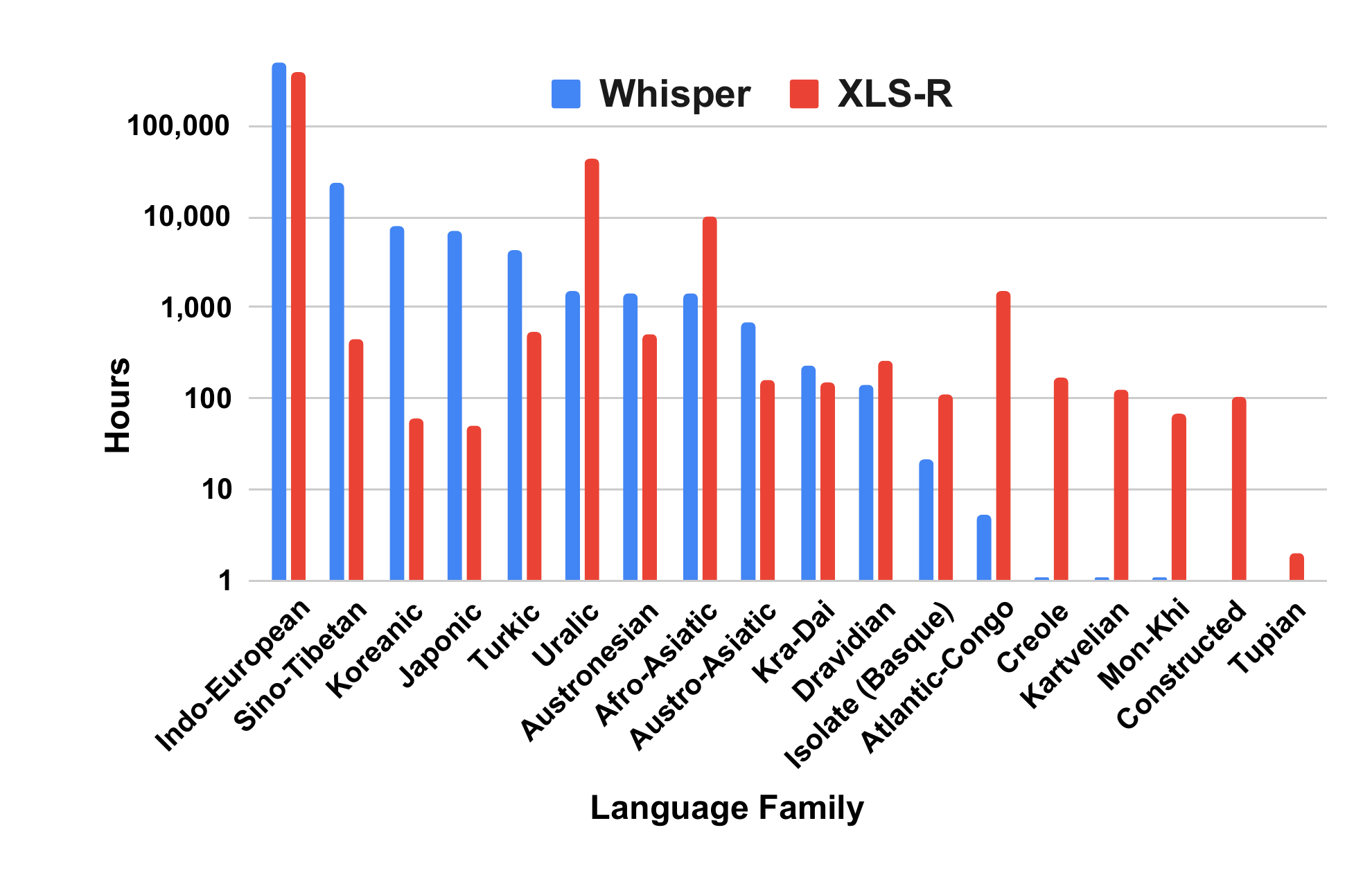}
    \caption{Hours of audio used to pre-train Whisper~\cite{radford2022robust} and XLS-R~\cite{babu2021xls} per language family.}
    \label{fig:hours}
\end{figure}

In this work, we compare XLS-R and Whisper for adaptation to languages unseen during pre-training.
We fine-tuned the models on 13 unseen languages in the FLEURS dataset~\cite{conneau2023fleurs}, on which neither model was pre-trained.
As shown in Figure~\ref{fig:hours}, the models have seen different amounts of languages during pre-training, and we discuss how this impacts the performance on the unseen languages.
We also fine-tuned on 18 seen languages in FLEURS and found that the number of hours seen per language during pre-training are predictive of how the models compare.

In summary, our contributions are: 1) We compare two state-of-the-art multilingual speech models with self-supervised and weakly-supervised pre-training strategies for adaptation to 13 unseen and 18 seen languages through fine-tuning; 2) We analyze the impact of the pre-training data and give suggestions for selecting either model for adaptation.
Our ArXiv paper version contains additional results in the Appendix.

\section{Methods}
Multilingual speech models have recently been trained on around 100 languages~\cite{adams-etal-2019-massively,Pratap2020,li2022massively,tjandra2022massively,lugosch2022pseudo,bapna2022mslam,zhang2023google,pratap2023scaling}.
However, few models are publicly available.
We use XLS-R~\cite{babu2021xls} and Whisper~\cite{radford2022robust} since they are publicly available, were trained on the most languages and hours of audio, and have different pre-training strategies.

\subsection{Self-Supervised Learning: XLS-R}
Self-Supervised Learning has proved to be effective for training ASR models.
Wav2Vec2.0~\cite{baevski2020wav2vec} was trained with large amounts of unlabeled audio and fine-tuned on smaller, labeled datasets to achieve good performance.
XLSR53, a multilingual version of Wav2Vec2.0, was trained on 50k hours of audio in 53 languages and fine-tuned for ASR in different languages.
Multilingual self-supervised training usually outperformed monolingual training, especially for lower-resource languages.
XLS-R~\cite{babu2021xls} scaled the pre-training to 436k hours of data in 128 languages, using publicly available datasets such as VoxPopuli~\cite{wang-etal-2021-voxpopuli}, CommonVoice~\cite{ardila-etal-2020-common}, Babel~\cite{gales2014speech}, and Multilingual LibriSpeech~\cite{pratap20_interspeech}.

\subsection{Weakly-Supervised Learning: Whisper}
Although Wav2Vec2.0 was pre-trained with unlabeled data which is easier to acquire than labeled data, it needs to be fine-tuned on each dataset with potentially different hyperparameters.
Radford~et~al.~\cite{radford2022robust} addressed this issue by developing Whisper, a supervised model trained on multilingual speech that can perform ``zero-shot'' ASR on different datasets (without fine-tuning).
The authors trained Whisper on 680k hours of speech from 99 languages.
In their experiments on English datasets, Whisper outperformed Wav2Vec2.0 without any fine-tuning.
Recent work also finds that Whisper can outperform other state-of-the-art ASR models when fine-tuned~\cite{gandhi2022esb}.

When tested on multilingual ASR, Whisper performed well on high-resource languages but performed poorly on languages it saw few hours of during pre-training.
We extend this analysis by comparing Whisper to XLS-R on multilingual ASR.
We also fine-tune Whisper, improving performance on seen languages and enabling adaptation to unseen languages.

\subsection{Language Adaptation}
Language adaptation aims to use representations learned from high-resource languages to improve performance on low-resource languages.
No audio in the target language is available in the zero acoustic resource setting~\cite{gales2014speech}. However, a more realistic setting is to assume a limited amount of transcribed audio.
Transferring representations from fully-supervised English models to other languages is a common setup~\cite{ardila-etal-2020-common,kessler2022adapter}, while others transfer from multilingual models. 
Early work transferred from models trained on three languages~\cite{grezl2014adaptation} while others scaled training to 100 languages~\cite{adams-etal-2019-massively}.
As self-supervised models became more common, cross-lingual transfer from methods like Contrastive Predictive Coding~\cite{oord2018representation} were also explored~\cite{riviere2020unsupervised}.
More recently, XLS-R has been shown to outperform Wav2Vec2.0 for multilingual cross-lingual transfer~\cite{khurana2022magic,krishna2021using,zhao2022improving}.

Several prior studies also fine-tune XLS-R and other self-supervised models on low-resource languages~\cite{pham22_interspeech,zhao2022improving}.
Our work differs since we focus on unseen languages and distinguish them from seen languages, we analyze how the pre-training data impacts the results, and we compare with Whisper.

Finally, we focus on adaptation to new languages and not on the \textit{language forgetting} phenomenon in the models.
Language forgetting could be studied as future work, with approaches including adaptors~\cite{kessler2022adapter,hou2021exploiting} and weight factorization~\cite{pham2022continually}.

\begin{table*}[htb]
\caption{Results on the FLEURS dataset. XLS-R 317M, Whisper-Small 244M, and Whisper-Medium 769M are fine-tuned on each language individually. CER=Character Error Rate ($\downarrow$ is better); PT=Pre-Train; FT=Fine-Tune; ZT=Zero-Shot (no fine-tuning).}
\vspace{-1em}
\label{tab:results}
	\centering
	\resizebox{0.89\linewidth}{!}{\begin{tabular}{lcccccccc}
		\toprule
        Language & Language & XLS-R & Whisper & XLS-R & Whisper-S & Whisper-M & Whisper-S & Whisper-M\\
        & Family & PT Hours & PT Hours & FT CER & FT CER & FT CER & ZT CER & ZT CER\\
        \midrule
        \bf{Seen Languages} \\
        \midrule
        \multicolumn{4}{l}{\it{XLS-R PT Hours $<$ Whisper PT Hours}} \\
        English & Indo-European & 69,493 & 438,218 & 6.45 & 3.63 & \bf{2.72} & 3.18 & \underline{2.81} \\
        Mandarin Chinese & Sino-Tibetan  & 90 & 23446 & 20.25 & \underline{9.55} & \bf{7.77} & 25.84 & 16.06 \\
        Korean & Koreanic & 61 & 7993 & 15.92 & 9.31 & \bf{5.27} & 7.56 & \underline{5.78} \\
        Japanese & Japonic & 49 & 7054 & 24.84 & 12.10 & \bf{7.22} & 14.47 & \underline{9.18} \\
        Indonesian & Austronesian & 41 & 1014 & 3.57 & 4.19 & \bf{2.85} & 6.83 & \underline{3.33} \\
        Arabic & Afro-Asiatic & 95 & 739 & \underline{5.99} & 6.86 & \bf{4.80} & 9.64 & 6.08 \\
        Ukrainian & Indo-European & 72 & 697 & 4.16 & 4.81 & \bf{3.20} & 6.39 & \underline{3.89} \\
        Vietnamese & Austro-Asiatic & 96 & 691 & 10.54 & 9.42 & \bf{6.42} & 11.05 & \underline{7.01} \\
        Thai & Kra-Dai & 20 & 226 & \underline{10.29} & 12.65 & \bf{8.82} & 24.05 & 17.75 \\
        \midrule
        \multicolumn{4}{l}{\it{XLS-R PT Hours $\approx$ Whisper PT Hours}} \\
        Azerbaijani & Turkic & 47 & 47 & \underline{5.84} & 8.05 & \bf{5.25} & 19.62 & 12.54 \\
        \midrule
        \multicolumn{4}{l}{\it{XLS-R PT Hours $>$ Whisper PT Hours}} \\
        Czech & Indo-European & 18514 & 192 & \underline{5.00} & 8.51 & \bf{4.69} & 11.30 & 6.43 \\
        Maltese & Afro-Asiatic & 9120 & 1.1 & \bf{4.08} & 8.80 & \underline{5.78} & 100.26 & 88.24 \\
        Bengali & Indo-European & 100 & 1.3 & \bf{6.04} & 11.08 & \underline{7.67} & 114.56 & 111.30 \\
        Swahili & Atlantic-Congo & 91 & 5.4 & \underline{4.88} & 6.91 & \bf{4.71} & 44.40 & 37.67 \\
        Afrikaans & Indo-European & 87 & 4.1 & \underline{9.31} & 13.83 & \bf{9.14} & 29.80 & 21.24 \\
        Hindi & Indo-European & 65 & 12 & \bf{5.94} & 8.21 & \underline{6.13} & 45.46 & 24.16 \\
        Khmer & Austro-Asiatic & 33 & 1.3 & \bf{12.49} & 27.18 & \underline{18.63} & 133.97 & 112.02 \\
        Burmese & Sino-Tibetan & 33 & 0.1 & \bf{11.58} & 30.46 & \underline{27.58} & 152.30 & 178.60 \\
        \multicolumn{4}{l}{\textit{Average CER for Seen Languages}} & \underline{9.29} & 10.86 & \bf{7.70}\\
        \midrule
        \bf{Unseen Languages} \\
        \midrule
        Asturian & Indo-European & 0 & 0 & \underline{5.29} & 7.09 & \bf{5.02} & N/A & N/A \\
        Kabuverdianu & Indo-European & 0 & 0 & \underline{4.65} & 5.64 & \bf{4.00} & N/A & N/A \\
        Sorani Kurdish & Indo-European & 0 & 0 & \bf{7.75} & 12.57 & \underline{9.89} & N/A & N/A \\
        Oromo & Afro-Asiatic & 0 & 0 & \bf{16.23} & 18.95 & \underline{17.57} & N/A & N/A \\
        Fula & Atlantic-Congo & 0 & 0 & \bf{16.07} & 20.87 & \underline{16.26} & N/A & N/A \\
        Kamba & Atlantic-Congo & 0 & 0 & \bf{12.87} & 19.20 & \underline{16.50} & N/A & N/A \\
        Sotho & Atlantic-Congo & 0 & 0 & \bf{7.35} & 13.90 & \underline{11.16} & N/A & N/A \\
        Nyanja & Atlantic-Congo & 0 & 0 & \bf{8.64} & \underline{13.37} & 14.69 & N/A & N/A \\
        Wolof & Atlantic-Congo & 0 & 0 & \bf{14.59} & 17.60 & \underline{14.71} & N/A & N/A \\
        Xhosa & Atlantic-Congo & 0 & 0 & \bf{6.34} & 12.48 & \underline{8.52} & N/A & N/A \\
        Igbo & Atlantic-Congo  & 0 & 0 & \bf{12.56} & \underline{19.34} & 19.64 & N/A & N/A \\
        Umbundu & Atlantic-Congo & 0 & 0 & \bf{18.13} & 24.69 & \underline{19.58} & N/A & N/A \\
        Luo & Nilo-Saharan  & 0 & 0 & \bf{5.96} & 8.55 & \underline{6.54} & N/A & N/A \\
        \multicolumn{4}{l}{\textit{Average CER for Unseen Languages}} & \bf{10.49} & 14.94 & \underline{12.62}\\
		\bottomrule
	\end{tabular}}
\end{table*}

\section{Experimental Setup}
\noindent \textbf{Dataset.}
We use the FLEURS~\cite{conneau2023fleurs} dataset, which covers 102 languages.
We selected this dataset since neither XLS-R nor Whisper was pre-trained on it,  there is a sufficient number of unseen languages, the data is approximately equal per language, and the data was verified by paid annotators.
The dataset contains around 2k sentences per language, and each sentence is spoken by up to 3 different speakers for a total of around 12 hours of audio per language.
The text was originally sourced from English Wikipedia and translated into other languages as the FLoRes-101 benchmark for machine translation~\cite{goyal-etal-2022-flores}.
For each language, we fine-tune the models on the training set, select the best checkpoint on the validation set, and report the performance on the test set.
We evaluate models with the Character Error Rate (CER) since languages such as Chinese, Japanese, Thai, and Burmese do not use regular spacing, which is consistent with prior work~\cite{ardila-etal-2020-common,conneau2023fleurs}.
We show the Word Error Rate (WER) results in section~\ref{sec:appendix-fleurs} of the Appendix.
We also show results on the CommonVoice dataset~\cite{ardila-etal-2020-common} in section~\ref{sec:appendix-commonvoice} of the Appendix.

\noindent \textbf{Languages.}
We used the 13 unseen languages in FLEURS which are common to XLS-R and Whisper.
We also used 18 seen languages.
We picked a roughly equal number of seen languages for which XLS-R saw more hours of audio during pre-training than Whisper, and vice-versa.
The languages use different scripts; the Latin script is used by 12 of the unseen languages and 8 of the seen languages.

\noindent \textbf{Model Sizes.}
XLS-R ranges in size from 24 layers / 317M parameters to 48 layers / 2.16B parameters, while Whisper ranges in size from 4 layers / 39M parameters to 32 layers / 1.55B parameters.
We compare XLS-R 317M with Whisper models with the closest number of parameters: 12 layers / 244M parameters (Whisper-Small) and 24 layers / 769M parameters (Whisper-Medium).
We show the results for XLS-R 965M and Whisper-Large-V2 in section~\ref{sec:appendix-fleurs} of the Appendix.

\noindent \textbf{Tokenization.}
For XLS-R, we used the language-specific characters from the FLEURS text as the outputs for ASR training.
For Whisper, we used the token IDs from the byte-level BPE tokenizer for ASR training, which can encode any UTF-8 text by converting characters to bytes.
This results in no out-of-vocabulary characters, although the encoding might not be efficient since one character of multiple bytes could be encoded as multiple tokens.
Whisper also expects a language token in the decoder input that is either predicted by its audio-based language ID system or manually specified. 
Since no token corresponds to the unseen languages, we tried setting the token equal to English or a language similar to each unseen language.
Both strategies resulted in similar results, so we used English as the language token for all unseen languages.

\noindent \textbf{Hyperparameters.}
XLS-R is fine-tuned with the CTC objective~\cite{graves2006connectionist}, and Whisper is fine-tuned with the Cross-Entropy objective.
We used Fairseq~\cite{ott2019fairseq}\footnote{\url{https://github.com/facebookresearch/fairseq/blob/main/examples/wav2vec/xlsr/README.md}} and HuggingFace~\cite{wolf-etal-2020-transformers}\footnote{\url{https://github.com/huggingface/community-events/tree/main/whisper-fine-tuning-event}} for fine-tuning XLS-R and Whisper, respectively, and adopted most of the default hyperparameters, with the following changes.
We fine-tuned XLS-R for 20k steps with a learning rate of 5e-5 and a gradient accumulation factor of 5.
We used a batch of around 1 minute of transcribed speech.
Unlike~\cite{babu2021xls}, we did not find it necessary to freeze the Transformer~\cite{vaswani2017attention} encoder's parameters at the beginning of training.
For Whisper, we used a batch of 48 recordings for the small model and 24 recordings for the medium model, and fine-tuned for 2k steps with a learning rate of 1e-5.
We used a single V100 32GB GPU, and each experiment took around 12 hours.

\noindent \textbf{Decoding.} 
We used greedy search for ASR inference with XLS-R.
We tried beam search without a language model (LM) but found no improvement.
We do not train LMs so that our results are independent of external modules used in the decoding process, and also since FLEURS contains little text to train them.
For inference with Whisper, we used greedy search for a fair comparison.
However, as discussed in Section~\ref{sec:unseen}, beam search decoding gives a slight improvement.

\noindent \textbf{Text Normalization.}
Since Whisper was trained with unnormalized transcripts, it generates transcripts with punctuation.
Hence, Radford~et~al.~\cite{radford2022robust} introduced a multilingual text normalizer to standardize both the Whisper transcripts and evaluation transcripts by removing punctuation.
The text in FLEURS was already normalized; however, we still found some punctuation and irrelevant characters.
Therefore, we applied the Whisper normalizer to all model-generated and ground-truth transcripts during evaluation to make a fair comparison between models with and without fine-tuning.
The Whisper normalizer erroneously introduces spaces to transcripts in a few languages; we discuss the impact of this in section~\ref{sec:appendix-spacing} of the Appendix.

\begin{figure*}[t]
    \centering
    \includegraphics[width=\linewidth]{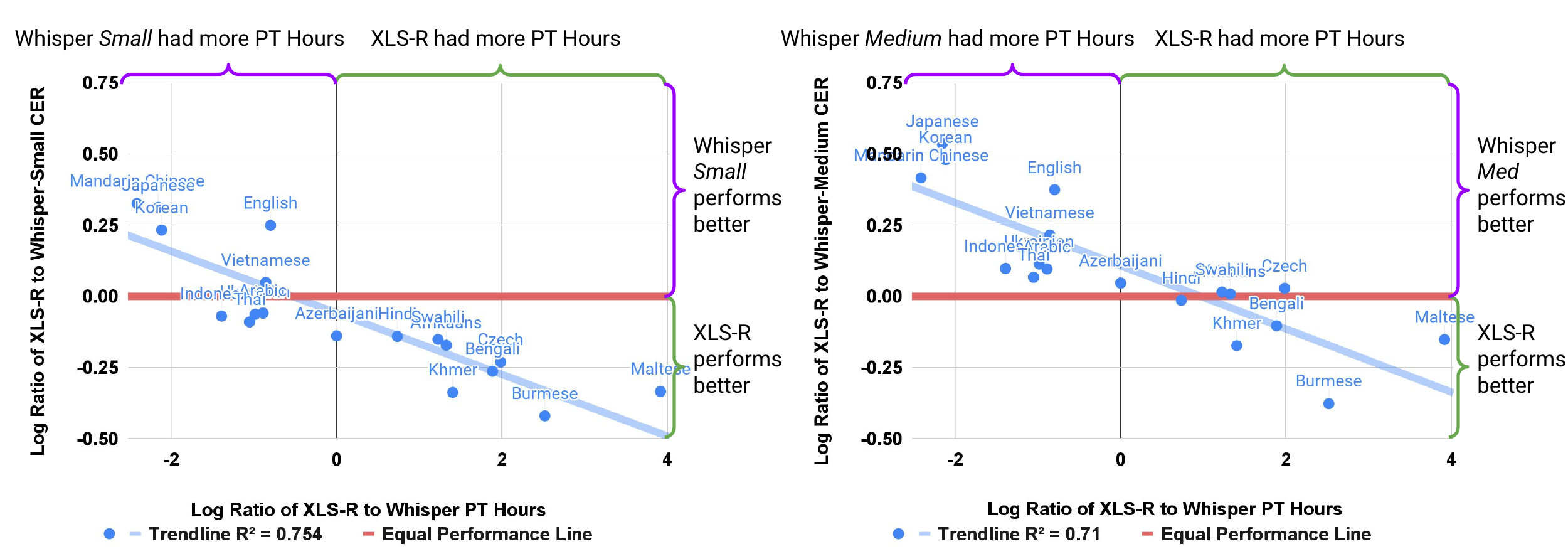}
    \caption{Comparing the performance of XLS-R 317M to Whisper Small 244M (left) and XLS-R 317M to Whisper Medium 769M (right) on seen languages in FLEURS. X-axis: $\log_{10}(\frac{\text{XLS-R \#PT Hours}}{\text{Whisper \#PT Hours}})$. Y-axis: $\log_{10}(\frac{\text{XLS-R CER}}{\text{Whisper CER}})$. }
    \label{fig:performance}
\end{figure*}

\section{Results}
\subsection{Seen Languages}
In Table~\ref{tab:results}, we show the results of fine-tuning XLS-R and Whisper on the seen and unseen languages in FLEURS.
We also show the Whisper results on seen languages without fine-tuning.
Both Whisper-Small and Whisper-Medium usually improved after fine-tuning.
The improvements are significant for languages with only a few hours of audio seen during pre-training; for example, CER improved from 88.24 to 5.78 for Whisper-Medium on Maltese (1.1 hours seen during pre-training).
However, for high-resource languages like English and Korean, the improvement was small, and the performance slightly decreased in some cases.
Whisper-Medium outperformed Whisper-Small on all seen languages both with and without fine-tuning.

Comparing XLS-R to Whisper, the results tend to depend on the number of hours seen per language during pre-training.
Whisper-Medium outperformed XLS-R both with and without fine-tuning on languages for which it saw more hours of audio during pre-training.
Whisper-Medium fine-tuned also outperformed XLS-R on some languages for which it saw fewer hours of audio during pre-training, such as Czech (4.69 vs 5.00) and Swahili (4.71 vs 4.88).
However, XLS-R outperformed Whisper-Medium on languages that it had seen far more of during pre-training, such as Hindi (5.94 vs 6.13) and Khmer (12.49 vs 18.63).
XLS-R outperformed Whisper-Small on most languages including those that Whisper had seen more hours of during pre-training, such as Indonesian (3.57 vs 4.19) and Arabic (5.99 vs 6.86). 
However, Whisper-Small outperformed XLS-R on languages that it had seen far more hours of during pre-training, such as Mandarin (9.55 vs 20.25) and Japanese (12.10 vs 24.84).

In Figure~\ref{fig:performance}, we compare the performance of XLS-R to Whisper-Small and XLS-R to Whisper-Medium based on the hours of pre-training data for each language.
The y-axis measures the relative CER and the x-axis measures the relative number of pre-training hours per language on a log-log scale.
Based on both plots, we observe a medium-to-strong correlation between the log of the relative CER and the log of the relative number of pre-training hours, which could be helpful for predicting whether XLS-R or Whisper will perform better on other seen languages.
The first line of best fit predicts that Whisper-Small will outperform XLS-R on languages for which it has seen at least 3.47 times more hours of audio than XLS-R during pre-training.
The second line of best fit predicts that XLS-R will outperform Whisper-Medium on languages for which it has seen at least 9.23 times more hours of audio during pre-training.
This result is intuitive since it predicts that a smaller model would outperform a larger model on languages for which it has been trained with far more hours of audio.
It is also interesting that the raw number of hours seen per language is predictive of the models' relative performance, despite the significant differences between XLS-R's self-supervised pre-training and Whisper's weakly supervised pre-training.

\subsection{Unseen Languages}
\label{sec:unseen}
The bottom half of Table~\ref{tab:results} shows the results of fine-tuning XLS-R and Whisper on the unseen languages.
We also attempted to test Whisper without fine-tuning but found that the transcripts were incomprehensible.
XLS-R outperformed Whisper-Small on all languages and outperformed Whisper-Medium on all languages except for Asturian and Kabuverdianu.
The models tend to perform worse on the unseen languages compared to the seen languages based on the average CER, but XLS-R's average CER only decreased by 12.9\% relative, while Whisper-Medium's average CER decreased by 63.9\% relative.

To understand the impact of the models' pre-training data on performance, we plotted the hours of audio seen during pre-training per language \textit{family} in Figure~\ref{fig:hours}.
Language families consist of languages that have a common ancestral language.
Whisper saw more audio in high-resource language families such as Indo-European (Whisper: 506k hours vs XLS-R: 379k hours), while XLS-R saw more audio in lower-resource language families such as Afro-Asiatic (Whisper: 1.4k hours versus XLS-R: 10k hours) and Atlantic Congo (Whisper: 5.4 hours versus XLS-R: 1.5k hours).
In our experiments, the model which saw more data in the corresponding language family usually performed better on the unseen language.
XLS-R did better on the Atlantic-Congo, Afro-Asiatic, and Nilo-Saharan languages, consistently outperforming the larger Whisper-Medium model.
Although neither model was pre-trained on Nilo-Saharan languages, the family is similar to Atlantic-Congo which explains XLS-R's better performance on Luo.
Whisper-Medium outperformed XLS-R on Indo-European languages, a family that Whisper had seen more data from, except for Sorani Kurdish.
Overall, these results show that the distribution of hours seen per language family can be useful to make predictions about how XLS-R and Whisper compare on unseen languages, despite the different pre-training methods.

For seen languages, the hours seen per language during pre-training appears to be a more important factor than the hours seen per family.
For example, XLS-R outperformed Whisper-Medium on languages that XLS-R had seen more hours of during pre-training, such as Hindi (Indo-European), Khmer (Austro-Asiatic), and Burmese (Sino-Tibetan), even though Whisper saw more hours in those families.
Similarly, XLS-R outperformed Whisper-Small on Indo-European languages that XLS-R had seen more hours of during pre-training, such as Czech, Bengali, and Afrikaans, even though Whisper saw more hours from the family in total.

Finally, we attempted to improve Whisper's performance on the unseen languages by using beam search instead of greedy search.
With a beam size of 5, Whisper's performance improved by around 3.34\% relative or 0.42 absolute on average, but the improvement was never large enough to outperform XLS-R.

\section{Conclusion}
We compared XLS-R and Whisper for adaptation to unseen languages not included during pre-training.
XLS-R largely outperformed Whisper on unseen languages, which we attribute to more data seen in the related language families during pre-training.
The hours per language seen during pre-training was predictive of how the models performed on seen languages.
Our results confirm that diverse pre-training data is essential for generalization and should ideally include more languages.
Since unlabeled audio is easier to collect, future self-supervised models like XLS-R may have greater potential to outperform supervised models like Whisper on other unseen languages, with the limitation that self-supervised models can't perform ASR without fine-tuning.
Future work could try ensembles of the two models, different tasks such as speech translation, or including other modalities such as vision~\cite{rouditchenko2023c2kd,rouditchenko21b_interspeech}.

\section{Acknowledgements}
This research was supported by the MIT-IBM Watson AI Lab, the MIT SuperCloud, and an NDSEG Fellowship to A.R.

\bibliographystyle{IEEEtran}
\bibliography{mybib}

\section{Appendix}

\subsection{Full Results on FLEURS}
\label{sec:appendix-fleurs}
We conducted additional experiments with XLS-R 965M and Whisper-Large-V2 1,550M. Table~\ref{tab:results_cer} shows the full CER results and Table~\ref{tab:results_wer} shows the full WER results.
Note that the CER and WER are the same for languages without regular spacing (Chinese, Japanese, Burmese, and Thai). 
For XLS-R 965M, we decreased the learning rate to 1e-5 and used a batch size of 50s of speech.
For Whisper Large-V2, we decreased the learning rate to 1e-6 and used a batch size of 4 samples due to GPU memory constraints. Future work should investigate parameter-efficient fine-tuning methods such as adaptors\footnote{\url{https://github.com/huggingface/peft}}.

Comparing the Whisper models without fine-tuning, Whisper-Large-V2 achieved the best CER and WER on all seen languages except for Vietnamese.
Note that our results without fine-tuning are slightly different than those reported in the original work~\cite{radford2022robust}, and one reason is that we used greedy decoding instead of beam search.
Whisper-Large-V2 typically performed better with fine-tuning than without, except on a few languages such as English and Indonesian.
After fine-tuning, Whisper-Large-V2 performed comparably to Whisper-Medium on seen languages; Whisper-Large-V2 obtained 2.5\% worse relative CER and 0.61\% better relative WER on average compared to Whisper-Medium.
On unseen languages, Whisper-Large-V2 performed noticeably worse compared to Whisper-Medium, obtaining 6.69\% worse relative CER and 2.7\% worse relative WER on average compared to Whisper-Medium.

XLS-R 965M tended to outperform XLS-R 317M. 
On seen languages, XLS-R 965M obtained 16.35\% better relative CER and WER on average compared to XLS-R 317M.
On unseen languages, the 965M model obtained 7.85\% better relative CER and 6.37\% better relative WER than the 317M model.

Comparing Whisper-Large-V2 and XLS-R 965M on seen languages, Whisper tended to outperform XLS-R on languages for which Whisper was trained with more hours, while XLS-R tended to outperform Whisper on languages for which XLS-R was trained with more hours.
On unseen languages, XLS-R outperformed Whisper on most languages except Kabuverdianu, an Indo-European language. 
The two models performed similarly on Asturian, another Indo-European language, although XLS-R slightly outperformed Whisper.
Overall, the trends closely follow those reported in the main paper, and the pre-training distribution of hours per language and language family are predictive of how the models compare.

\subsection{Results on CommonVoice}
\label{sec:appendix-commonvoice}
We conducted fine-tuning experiments on unseen languages in CommonVoice 11~\cite{ardila-etal-2020-common}. 
We converted the audio to mono and resampled it to 16 kHz. 
We applied the Whisper text normalizer to all training and test transcripts.
Training hyperparameters remained the same.
Results are shown in Table~\ref{tab:cv}. Surprisingly, XLS-R 317M outperformed both Whisper-Small and Whisper-Medium on most languages, except for Hill Mari and Saraiki.
We only expected XLS-R to outperform Whisper on Luganda, Meadow Mari, Hill Mari, and Erzya since XLS-R saw more hours in the Atlantic-Congo and Uralic families during pre-training.
However, XLS-R was pre-trained on other languages in CommonVoice while Whisper was not pre-trained on CommonVoice, therefore, XLS-R may have an advantage over Whisper since it was already trained on audio in the same domain.

\subsection{Spacing Introduced by the Whisper Normalizer}
\label{sec:appendix-spacing}
The Whisper multilingual normalizer has been found to introduce extra spacing in some languages\footnote{\url{https://github.com/openai/whisper/discussions/858}}.
In our experiments, this introduced additional spacing to the transcripts in Bengali, Khmer, Burmese, Thai, and Hindi from the FLEURS dataset, and the transcripts in Odia and Dhivehi from the CommonVoice dataset.
On these languages, we noticed that the CER results were only slightly lower (better) with the Whisper normalizer during evaluation, while the WER results were significantly lower (better) with the Whisper normalizer than without since the additional spaces makes predicting the correct words easier.
On the other languages, the results were similar both with and without the Whisper normalizer.
Since we applied the Whisper normalizer to all models, our comparisons between the models are fair.
However, the reported CER and WER for the languages previously mentioned are likely overstated.
Note that this issue impacts the original Whisper results~\cite{radford2022robust}, and the true WER for several languages could be worse than reported.

\begin{table*}[!ht]
\caption{Full CER results on the FLEURS dataset. XLS-R 317M, XLS-R 965M, Whisper-Small 244M, Whisper-Medium 769M, and Whisper-Large-V2 1,550M  are fine-tuned on each language individually. CER=Character Error Rate ($\downarrow$ is better); PT=Pre-Train; FT=Fine-Tune; ZT=Zero-Shot (no fine-tuning).}
\vspace{-1em}
\label{tab:results_cer}
\centering
\resizebox{\linewidth}{!}{\begin{tabular}{lccc|cc|ccc|ccc}
\toprule
        Language & Language & XLS-R & Whisper & X-317M & X-965M & Whisper-S & Whisper-M & Whisper-L & Whisper-S & Whisper-M & Whisper-L \\
        & Family & PT Hours & PT Hours & FT CER & FT CER & FT CER & FT CER & FT CER & ZT CER & ZT CER & ZT CER \\
        \midrule
        \bf{Seen Languages} \\
        \midrule
\multicolumn{2}{l}{\textit{XLS-R PT Hours $<$ Whisper PT Hours}} & & & & & & & & \\
English &Indo-European & 69,493 & 438,218 & 6.45 & 4.77 & 3.63 & 2.72 & \textbf{2.13} & 3.18 & 2.81 & \underline{2.14} \\
Mandarin Chinese &Sino-Tibetan  & 90 & 23446 & 20.25 & 16.71 & 9.55 & \underline{7.77} & \textbf{6.57} & 25.84 & 16.06 & 17.58 \\
Korean &Koreanic & 61 & 7993 & 15.92 & 12.46 & 9.31 & \underline{5.27} & \textbf{4.88} & 7.56 & 5.78 & 5.34 \\
Japanese &Japonic & 49 & 7054 & 24.84 & 21.65 & 12.10 & \underline{7.22} & \textbf{5.73} & 14.47 & 9.18 & 7.30 \\
Indonesian &Austronesian & 41 & 1014 & 3.57 & \underline{2.77} & 4.19 & 2.85 & 3.01 & 6.83 & 3.33 & \textbf{2.39} \\
Arabic &Afro-Asiatic & 95 & 739 & 5.99 & 5.18 & 6.86 & 4.80 & \textbf{3.95} & 9.64 & 6.08 & \underline{4.54} \\
Ukrainian &Indo-European & 72 & 697 & 4.16 & 3.31 & 4.81 & 3.20 & \textbf{1.98} & 6.39 & 3.89 & \underline{2.46} \\
Vietnamese &Austro-Asiatic & 96 & 691 & 10.54 & 11.18 & 9.42 & \underline{6.42} & \textbf{5.83} & 11.05 & 7.01 & 7.10 \\
Thai &Kra-Dai & 20 & 226 & 10.29 & 9.55 & 12.65 & \textbf{8.82} & \underline{8.92} & 24.05 & 17.75 & 12.82 \\
\midrule
\multicolumn{2}{l}{\textit{XLS-R PT Hours $\approx$ Whisper PT Hours}} & & & & & & & & \\
Azerbaijani &Turkic & 47 & 47 & 5.84 & \textbf{4.79} & 8.05 & 5.25 & \underline{5.24} & 19.62 & 12.54 & 6.51 \\
\midrule
\multicolumn{2}{l}{\textit{XLS-R PT Hours $>$ Whisper PT Hours}} & & & & & & & \\
Czech &Indo-European & 18514 & 192 & 5.00 & \textbf{3.40} & 8.51 & 4.69 & \underline{3.70} & 11.30 & 6.43 & 3.93 \\
Maltese &Afro-Asiatic & 9120 & 1.1 & \underline{4.08} & \textbf{3.22} & 8.80 & 5.78 & 7.67 & 100.26 & 88.24 & 28.22 \\
Bengali &Indo-European & 100 & 1.3 & \underline{6.04} & \textbf{5.28} & 11.08 & 7.67 & 9.34 & 114.56 & 111.30 & 74.13 \\
Swahili &Atlantic-Congo & 91 & 5.4 & 4.88 & \textbf{3.70} & 6.91 & \underline{4.71} & 5.24 & 44.40 & 37.67 & 11.78 \\
Afrikaans &Indo-European & 87 & 4.1 & 9.31 & \textbf{8.01} & 13.83 & 9.14 & \underline{8.15} & 29.80 & 21.24 & 14.03 \\
Hindi &Indo-European & 65 & 12 & \underline{5.94} & \textbf{4.84} & 8.21 & 6.13 & 11.83 & 45.46 & 24.16 & 13.00 \\
Khmer &Austro-Asiatic & 33 & 1.3 & \underline{12.49} & \textbf{12.15} & 27.18 & 18.63 & 24.14 & 133.97 & 112.02 & 110.09 \\
Burmese &Sino-Tibetan & 33 & 0.1 & \underline{11.58} & \textbf{9.78} & 30.46 & 27.58 & 30.96 & 152.30 & 178.60 & 114.95 \\
        \multicolumn{2}{l}{\textit{Average CER for Seen Languages}} & & & 9.29 & \underline{7.93} & 10.86 & \textbf{7.70} & 8.29 & 42.26 & 36.89 & 24.35 \\
        \midrule
        \bf{Unseen Languages} \\
        \midrule
Asturian & Indo-European & 0 & 0 & 5.29 & \textbf{4.23} & 7.09 & \underline{5.02} & 5.09 & N/A & N/A & N/A \\
Kabuverdianu & Indo-European & 0 & 0 & 4.65 & 4.45 & 5.64 & \underline{4.00} & \textbf{3.60} & N/A & N/A & N/A \\
Sorani Kurdish & Indo-European & 0 & 0 & \underline{7.75} & \textbf{7.17} & 12.57 & 9.89 & 9.99 & N/A & N/A & N/A \\
Oromo & Afro-Asiatic & 0 & 0 & \textbf{16.23} & \underline{16.49} & 18.95 & 17.57 & 17.55 & N/A & N/A & N/A \\
Fula & Atlantic-Congo & 0 & 0 & \underline{16.07} & \textbf{15.44} & 20.87 & 16.26 & 16.19 & N/A & N/A & N/A \\
Kamba & Atlantic-Congo & 0 & 0 & \underline{12.87} & \textbf{11.64} & 19.20 & 16.50 & 19.74 & N/A & N/A & N/A \\
Sotho & Atlantic-Congo & 0 & 0 & \underline{7.35} & \textbf{6.64} & 13.90 & 11.16 & 14.32 & N/A & N/A & N/A \\
Nyanja & Atlantic-Congo & 0 & 0 & \underline{8.64} & \textbf{6.83} & 13.37 & 14.69 & 13.82 & N/A & N/A & N/A \\
Wolof & Atlantic-Congo & 0 & 0 & \underline{14.59} & \textbf{13.22} & 17.60 & 14.71 & 15.71 & N/A & N/A & N/A \\
Xhosa & Atlantic-Congo & 0 & 0 & \underline{6.34} & \textbf{6.07} & 12.48 & 8.52 & 9.13 & N/A & N/A & N/A \\
Igbo & Atlantic-Congo  & 0 & 0 & \underline{12.56} & \textbf{11.70} & 19.34 & 19.64 & 22.81 & N/A & N/A & N/A \\
Umbundu & Atlantic-Congo & 0 & 0 & \underline{18.13} & \textbf{16.13} & 24.69 & 19.58 & 24.44 & N/A & N/A & N/A \\
Luo & Nilo-Saharan  & 0 & 0 & \underline{5.96} & \textbf{5.46} & 8.55 & 6.54 & 6.07 & N/A & N/A & N/A \\
        \multicolumn{2}{l}{\textit{Average CER for Unseen Languages}} & 	& & \underline{10.49} & \textbf{9.65} & 14.94 & 12.62 & 13.73 & & &\\
    \bottomrule
    \end{tabular}}
\end{table*}
\begin{table*}[!ht]
\caption{Full WER results on the FLEURS dataset. XLS-R 317M, XLS-R 965M, Whisper-Small 244M, Whisper-Medium 769M, and Whisper-Large-V2 1,550M  are fine-tuned on each language individually. WER=Word Error Rate ($\downarrow$ is better); PT=Pre-Train; FT=Fine-Tune; ZT=Zero-Shot (no fine-tuning).}
\vspace{-1em}
\label{tab:results_wer}
\centering
\resizebox{\linewidth}{!}{\begin{tabular}{lccc|cc|ccc|ccc}
\toprule
        Language & Language & XLS-R & Whisper & X-317M & X-965M & Whisper-S & Whisper-M & Whisper-L & Whisper-S & Whisper-M & Whisper-L \\
        & Family & PT Hours & PT Hours & FT WER & FT WER & FT WER & FT WER & FT WER & ZT WER & ZT WER & ZT WER \\
        \midrule
        \bf{Seen Languages} \\
        \midrule
\multicolumn{2}{l}{\textit{XLS-R PT Hours $<$ Whisper PT Hours}} & & & & & & & & \\
English & Indo-European & 69,493 & 438,218 & 18.91 & 14.43 & 8.65 & 6.51 & \underline{4.96} & 7.09 & 6.69 & \textbf{4.81} \\
Mandarin Chinese & Sino-Tibetan  & 90 & 23446 & 20.25 & 16.71 & 9.55 & \underline{7.77} & \textbf{6.57} & 25.84 & 16.06 & 17.58 \\
Korean & Koreanic & 61 & 7993 & 44.78 & 35.82 & 19.81 & \underline{13.44} & \textbf{11.67} & 20.07 & 16.04 & 14.32 \\
Japanese & Japonic & 49 & 7054 & 24.84 & 21.65 & 12.10 & \underline{7.22} & \textbf{5.73} & 14.47 & 9.18 & 7.3 \\
Indonesian & Austronesian & 41 & 1014 & 15.46 & 11.77 & 14.63 & 9.03 & \textbf{7.42} & 21.27 & 10.87 & \underline{7.43} \\
Arabic & Afro-Asiatic & 95 & 739 & 24.15 & 21.26 & 24.72 & 18.33 & \textbf{15.04} & 32.43 & 21.88 & \underline{17.14} \\
Ukrainian & Indo-European & 72 & 697 & 19.32 & 15.04 & 18.75 & 12.31 & \textbf{7.66} & 23.38 & 12.23 & \underline{8.04} \\
Vietnamese & Austro-Asiatic & 96 & 691 & 26.43 & 27.98 & 19.03 & 12.38 & \textbf{9.73} & 22.52 & 13.64 & \underline{12.34} \\
Thai & Kra-Dai & 20 & 226 & 31.32 & 28.82 & 12.65 & \textbf{8.82} & \underline{8.92} & 60.33 & 47.22 & 12.82 \\
\midrule
\multicolumn{2}{l}{\textit{XLS-R PT Hours $\approx$ Whisper PT Hours}} &  &  &  &  &  &  &  &  &  &  \\
Azerbaijani & Turkic & 47 & 47 & 26.19 & 21.70 & 30.65 & 20.38 & \textbf{20.10} & 64.34 & 41.15 & \underline{24.13} \\
\midrule
\multicolumn{2}{l}{\textit{XLS-R PT Hours $>$ Whisper PT Hours}} &  &  &  &  &  &  &  &  &  &  \\
Czech & Indo-European & 18514 & 192 & 21.08 & \underline{14.04} & 31.45 & 17.67 & \textbf{13.93} & 40.73 & 23.49 & 14.31 \\
Maltese & Afro-Asiatic & 9120 & 1.1 & \underline{15.33} & \textbf{11.98} & 29.23 & 19.57 & 25.39 & 121.10 & 118.38 & 78.08 \\
Bengali & Indo-European & 100 & 1.3 & \underline{12.63} & \textbf{11.20} & 21.04 & 14.24 & 17.84 & 133.33 & 113.82 & 106.04 \\
Swahili & Atlantic-Congo & 91 & 5.4 & 18.66 & \textbf{14.27} & 25.90 & \underline{18.46} & 21.00 & 97.72 & 97.77 & 43.48 \\
Afrikaans & Indo-European & 87 & 4.1 & 26.57 & \underline{22.68} & 34.67 & 24.04 & \textbf{22.06} & 68.03 & 52.67 & 38.11 \\
Hindi & Indo-European & 65 & 12 & 12.74 & \textbf{10.45} & 16.52 & \underline{11.83} & 20.84 & 63.54 & 35.94 & 22.99 \\
Khmer & Austro-Asiatic & 33 & 1.3 & \underline{24.50} & \textbf{23.75} & 48.83 & 34.13 & 44.18 & 173.21 & 121.37 & 108.29 \\
Burmese & Sino-Tibetan & 33 & 0.1 & \underline{21.07} & \textbf{17.43} & 30.46 & 27.58 & 30.96 & 200.54 & 136.99 & 114.95 \\
        \multicolumn{2}{l}{\textit{Average WER for Seen Languages}} & & & 22.46 & 18.94 & 22.70 & \textbf{15.76} & \underline{16.33} & 66.11 & 49.74 & 36.23 \\
        \midrule
        \bf{Unseen Languages} \\
        \midrule
Asturian & Indo-European & 0 & 0 & 18.26 & \textbf{14.32} & 23.06 & \underline{15.73} & 16.91 & N/A & N/A & N/A \\
Kabuverdianu & Indo-European & 0 & 0 & 16.76 & 15.98 & 19.27 & \underline{14.14} & \textbf{13.09} & N/A & N/A & N/A \\
Sorani Kurdish & Indo-European & 0 & 0 & \underline{34.76} & \textbf{32.07} & 48.81 & 39.35 & 38.30 & N/A & N/A & N/A \\
Oromo & Afro-Asiatic & 0 & 0 & \textbf{59.71} & \underline{61.31} & 70.54 & 63.06 & 64.17 & N/A & N/A & N/A \\
Fula & Atlantic-Congo & 0 & 0 & 49.92 & 48.69 & 55.88 & \textbf{47.89} & \underline{47.93} & N/A & N/A & N/A \\
Kamba & Atlantic-Congo & 0 & 0 & \underline{45.45} & \textbf{41.20} & 56.92 & 49.02 & 55.13 & N/A & N/A & N/A \\
Sotho & Atlantic-Congo & 0 & 0 & \underline{24.72} & \textbf{22.15} & 39.47 & 30.76 & 34.14 & N/A & N/A & N/A \\
Nyanja & Atlantic-Congo & 0 & 0 & \underline{38.91} & \textbf{31.23} & 49.05 & 48.76 & 43.13 & N/A & N/A & N/A \\
Wolof & Atlantic-Congo & 0 & 0 & 42.30 & \textbf{39.83} & 46.53 & \underline{40.88} & 41.82 & N/A & N/A & N/A \\
Xhosa & Atlantic-Congo & 0 & 0 & 33.03 & \textbf{32.67} & 52.57 & \underline{40.43} & 41.57 & N/A & N/A & N/A \\
Igbo & Atlantic-Congo  & 0 & 0 & \underline{37.66} & \textbf{36.62} & 47.69 & 45.68 & 51.97 & N/A & N/A & N/A \\
Umbundu & Atlantic-Congo & 0 & 0 & \underline{46.91} & \textbf{43.22} & 60.55 & 49.45 & 55.49 & N/A & N/A & N/A \\
Luo & Nilo-Saharan  & 0 & 0 & \underline{25.50} & \textbf{23.65} & 32.63 & 27.81 & 27.16 & N/A & N/A & N/A \\
        \multicolumn{2}{l}{\textit{Average WER for Unseen Languages}} & & & \underline{36.45} & \textbf{34.07} & 46.38 & 39.46 & 40.83 & & \\
    \bottomrule
    \end{tabular}}
\end{table*}
\begin{table*}[!ht]
\caption{Results on CommonVoice 11. XLS-R 317M, Whisper-Small 244M, and Whisper-Medium 769M are fine-tuned on each language individually. Hours indicates number of validated hours. CER=Character Error Rate, WER=Word Error Rate ($\downarrow$ is better).}
\vspace{-1em}
\label{tab:cv}
	\centering
	\resizebox{0.89\linewidth}{!}{\begin{tabular}      {lccccc|ccc}
		\toprule
        Language & Language & Hours & XLS-R & Whisper-S & Whisper-M & XLS-R & Whisper-S & Whisper-M\\
        & Family & & CER & CER & CER & WER & WER & WER \\
        \midrule
       
        \bf{Unseen Languages} \\
        \midrule
        Luganda & Atlantic-Congo & 409 & \textbf{4.47} & 7.93 & \underline{6.42} & \textbf{22.95} & 31.05 & \underline{24.74} \\
        Meadow Mari & Uralic & 134 & \textbf{3.05} & 4.63 & 	\underline{3.94} & 	\textbf{15.39} & 18.25 &	\underline{15.50} \\
        Uyghur & Turkic & 107 & \textbf{4.97} & 8.78 & \underline{5.88} & \textbf{25.50} & 38.50 & \underline{26.96} \\
        Central Kurdish & Indo-European & 105 & \textbf{5.83} & 9.82 & 7.90 & \textbf{30.37} & 43.45 & 35.11 \\
        Kurmanji Kurdish & Indo-European & 53 & \textbf{6.41} & 9.80 & \underline{7.43} & \textbf{26.62} & 36.77 & \underline{28.44} \\
        Dhivehi & Indo-European & 38 & \textbf{4.60} & 8.77 & \underline{7.47} & \textbf{6.43} & 12.61 & \underline{10.60} \\
        Hill Mari & Uralic & 17 & \underline{5.83} & \underline{5.83} & \textbf{4.19} & \textbf{18.69} & 	28.33 & \underline{20.62} \\
        Toki Pona & Constructed & 8 & \textbf{0.84} & 1.52 & \underline{1.16} & \textbf{3.06} & 5.44 & \underline{4.10} \\
        Saraiki & Indo-European & 3 & \underline{19.58} & 21.41 & \textbf{17.99} & \underline{56.03} & 60.63 & \textbf{53.05} \\
        Erzya & Uralic & 3 & \textbf{9.30} & 11.80 & \underline{9.60} & \textbf{43.63} & 54.92 & \underline{46.86} \\
        Odia & Indo-European  & 2 & \textbf{10.04} & 13.51 & \underline{12.89} & \textbf{23.41} & 30.66 & \underline{26.80} \\
        Sardinian & Indo-European & 2 & \textbf{11.93} & 22.85 & \underline{17.22} & \textbf{41.69} & 62.80 & \underline{48.95}\\
        \multicolumn{3}{l}{\textit{Average CER}} & \textbf{7.24} & 10.55 & \underline{8.51} & \textbf{26.15} & 35.28 & \underline{28.48} \\
		\bottomrule
	\end{tabular}}
\end{table*}

\end{document}